\documentclass[letterpaper, 10 pt, conference]{ieeeconf}  %

\IEEEoverridecommandlockouts                              %

\title{\LARGE \bf
Look Gauss, No Pose: Novel View Synthesis using Gaussian Splatting without Accurate Pose Initialization
}

\author{Christian Schmidt, Jens Piekenbrinck, Bastian Leibe%
\thanks{All authors are with the Computer Vision Group, RWTH Aachen University, \texttt{<lastname>@vision.rwth-aachen.de}.}
}

\usepackage{eccvabbrv}

\usepackage{amsfonts}
\usepackage{amsmath}
\usepackage{adjustbox}
\usepackage{bm}
\usepackage{booktabs}
\usepackage{xcolor}
\usepackage{multirow}

\graphicspath{figures}

\usepackage{animate}

\usepackage{hyperref}
\usepackage{url}

\newcommand{\RPEr}{$\text{RPE}_r$}
\newcommand{\RPEt}{$\text{RPE}_t$}
\newcommand{\thRPEr}{\RPEr$\downarrow$}
\newcommand{\thRPEt}{\RPEt$\downarrow$}
\newcommand{\thATE}{ATE$\downarrow$}
\newcommand{\thROT}{Rot.$\downarrow$}
\newcommand{\thPOS}{Trans.$\downarrow$}
\newcommand{\thPSNR}{PSNR$\uparrow$}
\newcommand{\thSSIM}{SSIM$\uparrow$}
\newcommand{\thLPIPS}{LPIPS$\downarrow$}

\newcommand{\TxSOthree}{\text{TxSO(3)}}
\newcommand{\SEthree}{\text{SE(3)}}
\newcommand{\sethree}{\text{se(3)}}
\newcommand{\Exp}{\operatorname{Exp}}
\newcommand{\Log}{\operatorname{Log}}

\newcommand{\mat}[1]{\mathbf{#1}}
\newcommand{\R}{\mathbb{R}}
\newcommand{\vect}[1]{\bm{#1}}

\newcommand{\PAR}[1]{\vskip4pt \noindent {\bf #1~}}
\newcommand{\PARbegin}[1]{\noindent {\bf #1~}}

\newcommand{\pderiv}[2]{\frac{\partial#1}{\partial#2}}
\newcommand{\grad}[1]{\pderiv{\mathcal{L}}{#1}}

\makeatletter
\g@addto@macro\normalsize{%
  \setlength\abovedisplayskip{4pt}
  \setlength\belowdisplayskip{3pt}
  \setlength\abovedisplayshortskip{4pt}
  \setlength\belowdisplayshortskip{3pt}
}
\makeatother

\usepackage[capitalize]{cleveref}
\crefname{section}{Sec.}{Secs.}
\Crefname{section}{Section}{Sections}
\crefname{table}{Tab.}{Tabs.}
\Crefname{table}{Table}{Tables}

\begin{document}

\maketitle
\thispagestyle{empty}
\pagestyle{empty}

\begin{abstract}
3D Gaussian Splatting has recently emerged as a powerful tool for fast and accurate novel-view synthesis from a set of posed input images.
However, like most novel-view synthesis approaches, it relies on accurate camera pose information, limiting its applicability in real-world scenarios where acquiring accurate camera poses can be challenging or even impossible.
We propose an extension to the 3D Gaussian Splatting framework by optimizing the extrinsic camera parameters with respect to photometric residuals.
We derive the analytical gradients and integrate their computation with the existing high-performance CUDA implementation.
This enables downstream tasks such as 6-DoF camera pose estimation as well as joint reconstruction and camera refinement.
In particular, we achieve rapid convergence and high accuracy for pose estimation on real-world scenes.
Our method enables fast reconstruction of 3D scenes without requiring accurate pose information by jointly optimizing geometry and camera poses, while achieving state-of-the-art results in novel-view synthesis.
Our approach is considerably faster to optimize than most competing methods, and several times faster in rendering.
We show results on real-world scenes and complex trajectories through simulated environments, achieving state-of-the-art results on LLFF while reducing runtime by two to four times compared to the most efficient competing method.
Source code will be available at \url{https://github.com/Schmiddo/noposegs}.
\end{abstract}

\section{Introduction}
\label{sec:intro}

Recently, novel-view synthesis (NVS) methods have emerged as a powerful tool in computer vision, enabling not only the generation of photo-realistic images from unseen viewpoints, but also downstream tasks such as dense 3D scene reconstruction~\cite{mildenhall2020nerf, mueller2022instant, kerbl3Dgaussians} and camera pose estimation relative to a trained model~\cite{lin2023icra:pnerf, yen2020inerf, fu2023cf3dgs}.
Solving these tasks benefits various applications, such as computer graphics, robotics, augmented and virtual reality, and 3D scene understanding.

However, most NVS methods depend on accurate camera pose information to build an initial scene representation, limiting the applicability of these approaches in real-world scenarios where acquiring reliable camera poses can be challenging or even impossible.

A promising line of research explores to perform camera pose estimation and 3D reconstruction jointly~\cite{lin2021barf, wang2021nerfmm, fu2023cf3dgs}, eliminating the need for perfect aligned camera poses or pose information at all.
Additionally, leveraging inherent dependencies between pose and scene information potentially leads to more robust estimations~\cite{lin2021barf, bian2023nope}.
Despite the significant advancements using novel-view synthesis methods, achieving robust pose estimation or joint reconstruction and pose estimation remain an ongoing challenge.

\begin{figure}
    \centering
    \includegraphics[width=0.95\linewidth]{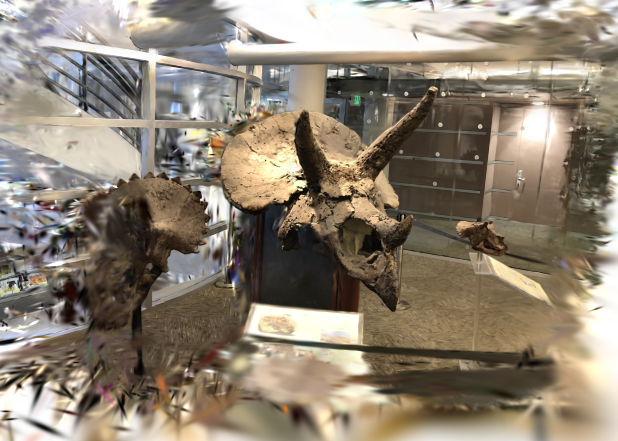}
    \caption{
    Our proposed approach enables fast, Gaussian Splatting based 3D reconstruction and photo-realistic novel-view synthesis while simultaneously estimating and refining the camera poses, as visualized on this reconstruction of the \emph{horns} scene from \emph{LLFF}~\cite{mildenhall2019llff}.
    This reconstruction was performed without camera pose information.
    }
    \label{fig:teaser}
\end{figure}

The most prominent methods for NVS are Neural Radiance Fields (NeRFs)~\cite{mildenhall2020nerf, mueller2022instant}, which achieve impressive results in terms of visual fidelity by modeling the radiance field as a continuous function using a neural network.
The scene can be rendered from any viewpoint using differentiable volumetric ray-marching, which involves shooting a ray for each pixel, querying the radiance field along the ray's path, and accumulating the information using $\alpha$-compositing.
NeRFs are optimized by a photometric reconstruction loss that compares the rendered pixels with the ground-truth, however due to the volumetric ray-marching they can be challenging to optimize and are computationally expensive in both training and rendering.

In contrast to NeRFs that represent a scene as a continuous implicit function, the recently proposed 3D Gaussian Splatting (3DGS)~\cite{kerbl3Dgaussians} employs an explicit scene representation based on a collection of anisotropic 3D Gaussians.
Combined with a differentiable tile-based splatting renderer implemented in CUDA, 3DGS offers highly efficient optimization even on commodity hardware while still delivering photo-realistic novel-view synthesis results in real-time.
However, similar to NeRFs, 3DGS requires accurate pose information for every training view.
In fact, the Gaussian Splatting algorithm is extremely sensitive to inaccurate camera poses, deteriorating even with very small distortions of the camera position and orientation.

Previous works in the NeRF literature perform pose estimation~\cite{yen2020inerf,lin2023icra:pnerf} as well as joint optimization of camera parameters and scene representation~\cite{lin2021barf, wang2021nerfmm, chng2022garf, Heo2023RobustCP} by propagating the gradients from the rendering loss back to the camera parameters.
Inspired by this, we propose to extend the 3D Gaussian Splatting framework through differentiable camera pose optimization. %
As we show, this optimization can directly be integrated into the CUDA rendering kernel, enabling very fast optimization. 

While similar in spirit to pose optimization in NeRFs, this approach faces additional challenges for 3DGS.
Theoretically, Gaussians have unlimited support; however, this support is usually cut off at some threshold, thus limiting the spatial range over which gradients can influence parts of the model.
On the other hand, highly anisotropic Gaussians allow the model to fit training views with high fidelity without accurate modeling of scene geometry.
We therefore introduce an anisotropy loss term to avoid overly fast convergence to suboptimal local minima and overfitting to training views.
Additionally, we improve the densification and pruning strategy of 3DGS to reduce shape-radiance ambiguities~\cite{zhang2020nerfpp}, resulting in improved geometry reconstruction and faster optimization and rendering speeds.

As our experiments show, the resulting approach is highly robust to noisy pose initialization, and can perform joint reconstruction and camera pose estimation with minimal assumptions about the input camera poses.
We show results for pose estimation as well as joint reconstruction and pose refinement on real-world scenes from the \emph{LLFF}~\cite{mildenhall2019llff} dataset, joint reconstruction and pose refinement on complex trajectories from \emph{Replica}~\cite{replica19arxiv}, and reconstruction without any known pose information on \emph{Tanks and Temples}~\cite{knapitsch2017tandt}.
In all evaluated cases, our proposed approach achieves state-of-the-art novel-view synthesis and pose estimation results, while enabling significantly faster runtimes.

In summary, our contributions are:
(1) We propose a novel differentiable camera pose estimation approach via Gaussian Splatting that lends itself to a highly efficient CUDA implementation.
(2) We improve the robustness of this approach to noisy pose initialization by proposing an anisotropy loss term.
(3) We demonstrate experimentally that our approach achieves state-of-the-art NVS and pose estimation / pose refinement results, while requiring only minimal assumptions on the input camera poses.

\section{Related Work}
\label{sec:related-work}

\PARbegin{Novel View Synthesis.}
Given a set of posed images of a scene, novel-view synthesis aims to generate realistic images from viewpoints that were not originally captured.
In the past years, methods have focused on representing the scene implicitly as radiance field that can be rendered using volumetric ray-marching~\cite{mildenhall2020nerf, chen2022tensorf, mueller2022instant}.
These approaches often combine some positional encoding together with an MLP, collectively referred to as Neural Radiance Fields (NeRFs).
Each pixel of the image is computed as the approximated integral over a ray through the NeRF volume, which usually requires many evaluations of the radiance field, making these approaches very slow to train and evaluate.
Follow-up methods improved the speed and quality of NeRFs, leading to state-of-the-art methods for novel-view synthesis.
These speedups were possible by leveraging spatial data structures with the combination of smaller neural networks \cite{chen2022tensorf, xu2022point, mueller2022instant} or even no neural networks \cite{fridovich2022plenoxels}.

Compared to the aforementioned NeRF methods that rely on ray-casting, point-based techniques use an explicit unordered set of geometry as representation of the scene, \eg point clouds, which is rendered using differentiable rasterization or splatting algorithms~\cite{zwicker2001ewavolume, zwicker2001surface, ren2002object, lassner2021pulsar}.
Kerbl~\etal~\cite{kerbl3Dgaussians} recently proposed 3D Gaussian Splatting (3DGS), a novel point-based method, that utilizes a set of anisotropic Gaussians to represent the scene.
A key advantage of 3DGS is its efficient rendering achieved through a differentiable tile-based splatting algorithm that is implemented in CUDA, enabling fast optimization and real-time rendering.

\PAR{Camera Pose Estimation.}
Estimating camera poses from images is a prevalent task in computer vision.
Existing pose estimation methods typically rely on a known 3D scene to establish 2D-3D correspondences that are used to derive the poses.
In recent years, research has explored pose estimation based on the implicit 3D representation of NeRFs.
Some approaches have adopted the idea of correspondences and combined them with NeRFs \cite{avraham2022nerfels, li2023nerfpose}.
Another line of work leverages the capabilities of photo-realistic rendering from any viewpoint for a direct comparison with the actual image in order to estimate the camera pose~\cite{yen2020inerf, lin2023icra:pnerf}.
However, NeRFs can be challenging to optimize and are computationally expensive to train and render~\cite{yen2020inerf,lin2021barf,bian2023nope}.

Recently, iComMa~\cite{sun2023icomma} is among the first to perform camera pose estimation based on 3DGS.
Compared to our work, they use an additional local feature matching model for an end-to-end matching loss, whereas our method is based on the photometric loss.

\PAR{Joint Reconstruction and Pose Estimation.}
The aforementioned novel-view synthesis techniques rely on posed images to densely reconstruct the 3D scene.
Consequently, Structure from Motion (SfM) methods, \eg COLMAP~\cite{schonberger2016colmap}, are frequently employed to estimate camera poses and acquire an initialization for point-based approaches.
Despite their prevalence, SfM methods suffer from inaccurate camera pose estimations due to unreliable keypoint matches in regions with low texture or repetitive patterns.
We demonstrate in \cref{tab:ablation_vanilla_3dgs} that our method with pose refinement achieves comparable or better novel-view synthesis results compared to relying solely on SfM initialization.

For NeRFs, there have been several works that tackle the task of jointly reconstructing 3D geometry and estimating camera poses~\cite{wang2021nerfmm, lin2021barf, chng2022garf, bian2023nope, Heo2023RobustCP, liu2023baangp, cheng2024JointTensoRF}.
Notably, Nope-NeRF~\cite{bian2023nope} showed that reconstructing a scene without prior assumptions on camera poses is possible, using only geometric cues from a monocular depth estimator.

Concurrent to our work, Colmap-Free Gaussian Splatting (CF-3DGS)~\cite{fu2023cf3dgs} proposes to estimate the relative poses between consecutive video frames and iteratively extend the set of Gaussians.
While we experiment with a similar initialization scheme for the trajectory, we use a different process to build a consistent Gaussian representation.
Notably, our method does not require an iterative estimation of poses and therefore has lower runtime, especially for long video sequences.
Additionally, our method can be adjusted to work on unordered image collections, which to the best of our knowledge makes us the first to support this for 3DGS.
Consequently, the runtime of our method is not linear in the number of input frames.
We compare against CF-3DGS in section~\cref{ssec:comparison-sota}.

\section{Method}
\label{sec:method}

We first recapitulate the original differentiable Gaussian Splatting approach (\cref{sec:splatting}).
We then describe our extension to this formulation in~\cref{sec:pose-gradients}, and details for the proposed optimization procedure in~\cref{sec:method:pose-refinement,sec:relative-pose-estimation}.

\subsection{Gaussian Splatting}
\label{sec:splatting}
Gaussian Splatting~\cite{kerbl3Dgaussians} represents a 3D scene as a set of oriented anisotropic Gaussians $\mathcal{G}$.
Every Gaussian is parametrized by a 3D mean $\vect{\mu}_i$, a covariance matrix $\mat{\Sigma}_i$, a set of feature vectors $\vect{f}_i$ and a scalar opacity $o_i$.
We denote the corresponding 2D quantities, \ie projected mean and covariance, with $\vect{\hat\mu}_i$ and $\mat{\hat\Sigma}_i$.
In order to facilitate gradient based optimization, the 3D covariance matrix is decomposed into an orientation represented as a unit quaternion $\vect{q}_i$ and a scale vector $\vect{s}_i$ such that
\begin{equation}
    \mat{\Sigma}_i = \mat{R}(\vect{q}_i)\text{diag}(\vect{s}_i)\text{diag}(\vect{s}_i)^T\mat{R}(\vect{q}_i)^T.
\end{equation}
To render a view from a given camera with intrinsics $\mat{K}$ and extrinsics $\mat{T} = (\mat{R}_c, \vect{t}_c)$, the Gaussians are projected onto the image plane and rasterized via alpha blending.
More formally, the projected 2D mean of a Gaussian is $\vect{\hat\mu}_i = \pi(\mat{K}\mat{T}\vect{\mu}_i)$ where $\pi$ is the projective transformation $\pi(\vect{p}) = (p_x/p_z, p_y/p_z, 1)$.
The 2D covariance $\mat{\hat\Sigma}$ of a Gaussian as projected onto the viewing plane of camera $c$ is then computed based on a first-order Taylor approximation of $\pi$~\cite{kerbl3Dgaussians},
\begin{equation}
    \mat{\hat\Sigma}_i = [\mat{J}_i\mat{R}_c\mat{\Sigma}_i\mat{R}_c^T\mat{J}_i^T]_{2\times 2},
\end{equation}
where $\mat{J}_i$ is the Jacobian of the aforementioned approximation, and $[.]_{2\times 2}$ denotes the upper two-by-two submatrix.

The influence of a splat $(\vect{\hat\mu}_i, \mat{\hat\Sigma}_i)$ on a pixel $\vect{p}$ is determined by the 2D distribution of the splat and its opacity $o_i$:
\begin{equation}
    \alpha_i = o_i \exp\left(-\frac{1}{2} (\vect{p} - \vect{\hat\mu}_i) \mat{\hat\Sigma}_i (\vect{p} - \vect{\hat\mu}_i)\right).
\end{equation}
The color of a splat is modeled via spherical harmonics $\vect{c}_i = \text{SH}(\vect{f}_i, \vect{d}_i^c)$, which calculate view-dependent color using a set of basis functions, learnable parameters $\vect{f}_i$, and the viewing direction $\vect{d}_i = \frac{\vect{\mu}_i - \vect{o}_c}{||\vect{\mu}_i - \vect{o}_c||}$.
Note that $\vect{o}_c = -\mat{R}_c^{-1} \vect{t}_c$ denotes the position of the camera center in world coordinates.

The splats are then sorted by depth and are composited front-to-back via alpha blending in order to find the final pixel color $\mat{C}_{xy}$:
\begin{align}
    \mat{C}_{xy}=\sum_{i}^{N} \vect{c}_{i} \alpha_{i} \prod_{j}^{i-1}(1-\alpha_{j}).
\end{align}
The term $T_i = \alpha_i\prod_j^{i-1}(1 - \alpha_j)$ is also called \emph{transmittance}.

The differentiable rasterizer of Kerbl~\etal~\cite{kerbl3Dgaussians} efficiently computes gradients for positions, orientations, scalings, opacities, and spherical harmonics of all Gaussians, allowing rapid fitting of 3D scenes given a few posed images.
However, it does not provide gradients for the camera extrinsics~$\mat{T}$.

\subsection{Camera pose optimization via Gaussian Splatting}
\label{sec:pose-gradients}
We propose to extend the Gaussian Splatting formulation to also compute gradients for the extrinsic camera parameters $\mat{R}_c$ and $\vect{t}_c$.
To this end, we model the camera pose as an element of $\SEthree$, the Lie group of 3D rigid body motions, and derive the gradients for group elements in the respective tangent space.
This allows us to use the renderer together with existing libraries for tangent space backpropagation, enabling fast and accurate pose estimation and other downstream tasks.

A Lie group is a group that is also a smooth manifold.
The lie algebra of the group $\SEthree$ is denoted $\sethree$ and is defined as the tangent space at the identity element; it is isomorphic to $\R^6$.
We use the hat $^\wedge :\R^6 \rightarrow \sethree$ and vee $^\vee : \sethree \rightarrow\R^6$ operators to map between $\R^6$ and $\sethree$.
The \emph{exponential map} $\Exp(\tau): \R^6 \rightarrow \SEthree$ and its counterpart, the \emph{logarithmic map} $\Log(\mathcal{X}): \SEthree\rightarrow\R^6$ map elements of $\R^6$ to elements of the Lie group and vice versa.
We can then parametrize the camera transform as $\mat{T} = \Exp(\vect{\tau}), \vect{\tau}\in\R^6$.
For simplicity of notation, we still denote gradients on $\vect{\tau}$ as $\grad{\mat{T}}$.

\PAR{Lie Groups.}
Given a function $f:\mathcal{M}\rightarrow\mathcal{M}^\prime$ that maps between two Lie groups $\mathcal{M}$ and $\mathcal{M}^\prime$, it is possible to extend the differential to compute gradients with respect to group elements~\cite{sola2018microlie}:
\begin{equation}
    \text{D}f(\mathcal{X})[\vect{v}] = \underset{t\rightarrow0}{\lim} \frac{\Log(f(\Exp(t\vect{v})\mathcal{X}) f(\mathcal{X})^{-1})}{t},
\end{equation}
where $\vect{v}$ is an element of the tangent space of $\mathcal{X}$.
Intuitively, $\vect{v}$ denotes the direction of the differential.
This allows to extend the notion of Jacobians as
\begin{equation}
    \label{eq:lie-jacobian}
    \left(\pderiv{f(\mathcal{X})}{\mathcal{X}}\right)_{ij} = \underset{t\rightarrow0}{\lim} \frac{<\Log(f(\Exp(t\vect{e}_j)\mathcal{X}) f(\mathcal{X})^{-1}), \vect{e}_i^\prime>}{t},
\end{equation}
where $\vect{e}_i,\vect{e}_j^\prime$ are orthonormal basis vectors of the input and output spaces of $f$, respectively, and $<\!\!\cdot,\cdot\!\!>$ denotes the vector product operator.
Note that every Euclidean space, together with the usual vector addition, forms a Lie group with its Lie algebra being itself.
Thus~\cref{eq:lie-jacobian} generalizes the usual notion of Jacobians to mappings between arbitrary Lie groups.

\PAR{Deriving the Jacobians.}
During optimization, our aim is to minimize some loss function $\mathcal{L}$ using gradient descent.
For this, we need to determine the gradient of the loss with respect to the camera parameters, $\grad{\mat{T}}$.
Three terms contribute to $\grad{\mat{T}}$: the 2D mean $\vect{\hat\mu}_i$, the 2D covariance matrices $\mat{\hat\Sigma}_i$,  and the view-dependent color $\vect{c}_i = \text{SH}(\vect{f}_i, \vect{d}_i^c)$.
Using the chain rule, we get
\begin{multline}
    \grad{\mat{T}} = \sum_{i}
        \grad{\vect{\hat\mu}_i}\pderiv{\vect{\hat\mu}_i}{\vect{\mu}_i^c}\pderiv{\vect{\mu}^c_i}{\mat{T}} \\
        + \grad{\mat{\hat\Sigma}_i} \left(
            \pderiv{\mat{\hat\Sigma}_i}{\mat{J}_i}\pderiv{\mat{J}_i}{\vect{\mu}_i^c}\pderiv{\vect{\mu}_i^c}{\mat{T}}
            + \pderiv{\mat{\hat\Sigma}_i}{\mat{R}_c}\pderiv{\mat{R}_c}{\mat{T}}
        \right)
        + \grad{\vect{c}_i}\pderiv{\vect{c}_i}{\vect{o}_c}\pderiv{\vect{o}_c}{\mat{T}},
\end{multline}
where we use $\vect{\mu}_i^c = \mat{T}\vect{\mu}_i$ to denote the mean of Gaussian $i$ in the reference frame of camera $c$.
The original formulation already computes $\grad{\vect{\mu}_i^c}, \grad{\mat{\hat\Sigma}_i}, \pderiv{\mat{\hat\Sigma}_i}{\vect{\mu}_i^c}$, and $\grad{\vect{c}_i}$.

The Jacobian of a transformation applied to a point $\vect{y} = \mat{T}\vect{x}$ is well-known~\cite{sola2018microlie} as
\begin{equation}
    \pderiv{\vect{y}}{\mat{T}} = \left(\mat{I}\,|\,-[\vect{y}]^\times\right),
\end{equation}
where $[\vect{y}]^\times$ denotes the 3-by-3 skew matrix with entries of $\vect{y}$, and $(\cdot |\cdot)$ denotes matrix concatenation; $\pderiv{\vect{y}}{\mat{T}}$ is a 3-by-6 matrix relating changes in $\vect{y}$ with changes in the tangent space representation of $\mat{T}$.
The Jacobian of the group inverse on $\SEthree$ can be found as~\cite{teed2021tangent}
\begin{equation}
    \pderiv{\mat{T^{-1}}}{\mat{T}} = -\begin{pmatrix}
        \mat{R} & [\vect{t}]^\times \mat{R} \\
        \vect{0} & \mat{R}
    \end{pmatrix},\quad\mat{T} = (\mat{R}, \vect{t}).
\end{equation}
With this, we can compute the gradient contributions of the 2D/3D mean and the spherical harmonics:
\begin{align}
    \pderiv{\vect{\mu}^c_i}{\mat{T}} &= \left( \mat{I}\,|\,-[\vect{\mu}_i^c]^\times \right) \\
    \pderiv{\vect{c}_i}{\mat{T}} &= \pderiv{\vect{c}_i}{\vect{o}_c}\pderiv{\vect{o}_c}{\mat{T}} = \pderiv{\vect{c}_i}{\vect{\mu}_i} (\mat{R}_c \,|\, [\vect{o}_c]^\times \mat{R}_c),
\end{align}
where we used the fact that spherical harmonics are only dependent on the direction from 3D mean $\vect{\mu}_i$ to the camera center $\vect{o}_c$, and $\vect{o}_c = -\mat{R}_c^{-1} \vect{t}_c$.

To compute the gradient contribution of the term $\mat{\hat\Sigma}_i = \mat{J}_i\mat{R}_c\mat{\Sigma}_i\mat{R}_c^T\mat{J}_i^T$, we apply~\cref{eq:lie-jacobian} on $f(\mat{T}) = \mat{R}_c$.
Here, $f$ is a function mapping from $\SEthree$ to $\R^{3\times 3}$, thus the Jacobian can be written as a matrix in $\R^{3\times 9}$, relating entries of $\vect{\tau}$ with entries of $\mat{R}_c$:
\begin{align}
    \pderiv{f(\mat{T})}{\mat{T}} &= \pderiv{}{\vect{\tau}}|_{\vect{\tau}=0} \Exp(\vect{\tau}) \mat{R}_c \\
    &= ([\vect{e}_1]^\times \mat{R}_c \,|\, [\vect{e}_2]^\times \mat{R}_c \,|\, [\vect{e}_3]^\times \mat{R}_c ).
\end{align}
We integrate these computations with the existing Gaussian Splatting rasterizer using the CUDA programming environment.
This allows us to compute accurate gradients on the camera parameters with minimal overhead, leading to significant reductions of runtime compared to other approaches.

\subsection{Splatting with Pose Optimization}
\label{sec:method:pose-refinement}
For joint reconstruction and pose refinement, we iteratively optimize over all Gaussian parameters and all camera poses.
Given a set of initial pose estimates $\mathcal{P} = \{\mat{T}_1, \ldots, \mat{T}_K\}$ corresponding to training images $I_1, \ldots, I_K$, and an initial set of Gaussians $\mathcal{G}$, we want to find optimal poses $\mathcal{P}^*$ and an optimal scene representation $\mathcal{G}^*$ such that
\begin{equation}
    \mathcal{P}^*, \mathcal{G}^* = \underset{\mathcal{P}, \mathcal{G}}{\arg\min} \left( \sum_{I} \mathcal{L}(\hat I, I)\right).
\end{equation}
Our loss $\mathcal{L}$ is a combination of an image-based loss $\mathcal{L}_{\text{RGB}}$ and an additional 3D anisotropy loss term $\mathcal{L}_{\text{An}}$ to regularize the scene representation:
\begin{equation}
    \mathcal{L} = \mathcal{L}_{\text{RGB}} + \mathcal{L}_{\text{An}}.
\end{equation}
The image-based loss $\mathcal{L}_{\text{RGB}}$ is a weighted sum of an $L_1$ loss and a DSSIM term, as in the original Gaussian Splatting formulation~\cite{kerbl3Dgaussians}:
\begin{align}
    \mathcal{L}_{\text{RGB}} = (1 - \beta) \mathcal{L}_{L_1} + \beta \mathcal{L}_{DSSIM}~,
\end{align}
where we use $\beta = 0.2$.
The regularization term $\mathcal{L}_{\text{An}}$ helps avoiding overly fast convergence of the Gaussians to suboptimal local minima and overfitting to the training views.
It limits the ratio between the major and minor axis of each Gaussian:
\begin{equation}
    \mathcal{L}_{\text{An}} = \frac{1}{N} \sum_{i=1}^{N} [\max \{\vect{s}_i\} / \min \{\vect{s}_i\} - r]_+,
\end{equation}
ensuring that Gaussians do not degenerate into arbitrarily thin spikes.

Additionally, we employ an adaptive thresholding scheme to remove transparent Gaussians.
The default densification and pruning strategy in 3DGS leads to a large number of semi-transparent Gaussians which contribute little to the actual scene representation while increasing shape-radiance ambiguity~\cite{zhang2020nerfpp}.
We adapt the opacity threshold so that at most $N_{\text{tgt}}$ Gaussians remain after pruning.
To make this scheme more effective, we add an $L_1$ loss on the opacity of each Gaussian during the first 10k steps.
The remaining Gaussians cover larger areas and are more opaque, leading to reduced shape-radiance ambiguities, improved geometry reconstruction, and faster rendering.

\subsection{Camera pose estimation}
\label{sec:relative-pose-estimation}
While the focus of our method lies on \emph{pose refinement}, we can also use it to \emph{estimate} an unknown camera pose, \eg, from a trained model or between nearby frames in a video.
This allows us to reconstruct scenes with just minimal assumptions on the camera pose distribution. (Essentially, we only require that the cameras show partially overlapping geometry). We show results on videos from the Tanks and Temples dataset~\cite{knapitsch2017tandt} in~\cref{ssec:comparison-sota}.
The idea here is to estimate a rough trajectory from neighbouring frames and then to refine this trajectory with our method for joint reconstruction and pose refinement.

For two nearby frames $I_t, I_{t+1}$ of an input video stream, we can estimate the relative pose between them as follows.
Using a pretrained off-the-shelf monocular depth estimator $D$, we compute a relative depth map $D(I_t)$.
We then use this depth map and the known camera intrinsics $\mat{K}$ to unproject a subset of pixels into a point cloud $P_t$.
We interpret these points as isotropic Gaussians and set their color to the respective pixel color and their scale to the mean 3D distance to the three nearest neighbours in the point cloud, yielding a set of 3D Gaussians $\mathcal{G}_t$.
We can then optimize $\mathcal{G}_t$ to better depict the input frame:
\begin{equation}
    \mathcal{G}_t^* = \underset{\mathcal{G}}{\arg\min} \left(\mathcal{L}_{\text{RGB}}(\hat I_t, I_t)\right).
\end{equation}
This optimization converges very quickly (typically in less than 100 steps).
Due to the efficient rasterizer and the limited number of 3D points, this initialization takes less than a second per frame.
After initializing $\mathcal{G}_t$, we estimate the local transformation $\mat{T}_{t+1}$ between frames $I_t$ and $I_{t+1}$ such that it minimizes the rendering loss of $\mathcal{G}_t$ compared to frame $I_{t+1}$:
\begin{align}
    \mat{T}_{t}^{*} = \underset{\mat{T}}{\arg\min}\ \mathcal{L}_{\text{RGB}}^M(\hat I_t, I_{t+1}).
\end{align}
Here, we use a confidence-masked loss function $\mathcal{L}_{\text{RGB}}^M$.
In particular, we apply a mask $M$ on the rendered image before computing the $L_1$ loss for pose estimation in order to evaluate the loss only on pixels with high accumulated alpha.
This makes the optimization more robust to moving objects and gaps in the rendering due to missing geometry.
We define the mask as a pixel-wise threshold on the accumulated transmittance:
\begin{equation}
    M_{x,y} = \begin{cases}
        1 & \text{if } \sum_i^N T_i > 0.99 \\
        0 & \text{otherwise}.
    \end{cases}
\end{equation}
The masked loss is then
\begin{equation}
    \mathcal{L}_{L_1}^M(\hat I, I) = \sum_{x,y} |M_{x,y}(I_{x,y} - \hat I_{x,y})|~.
\end{equation}

\section{Experiments}
\label{sec:experiments}

We perform experiments on three tasks: \emph{camera pose estimation} (\cref{sec:pose-estimation}), \emph{joint reconstruction and pose refinement} (\cref{sec:pose-refinement}), and \emph{reconstruction without pose information} (\cref{ssec:comparison-sota}).

\begin{figure*}
    \centering
    \vspace{0.2cm}
    {
    \setlength\tabcolsep{1.5pt}
    \begin{tabular}{c cccccc}
         & GT & BARF~\cite{lin2021barf} & GARF~\cite{chng2022garf} & MRHE~\cite{Heo2023RobustCP} & JTRF~\cite{cheng2024JointTensoRF} & Ours \\
         \rotatebox{90}{\makebox[2.7cm]{flower}} &
            \includegraphics[width=2.7cm]{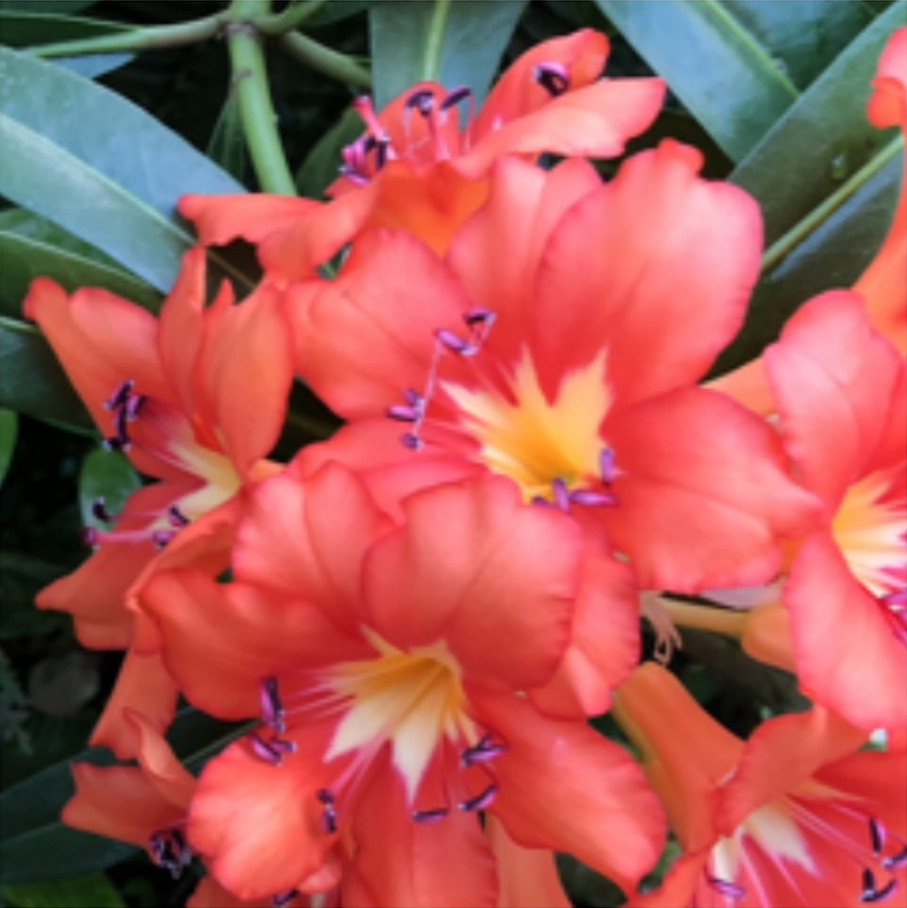} &
            \includegraphics[width=2.7cm]{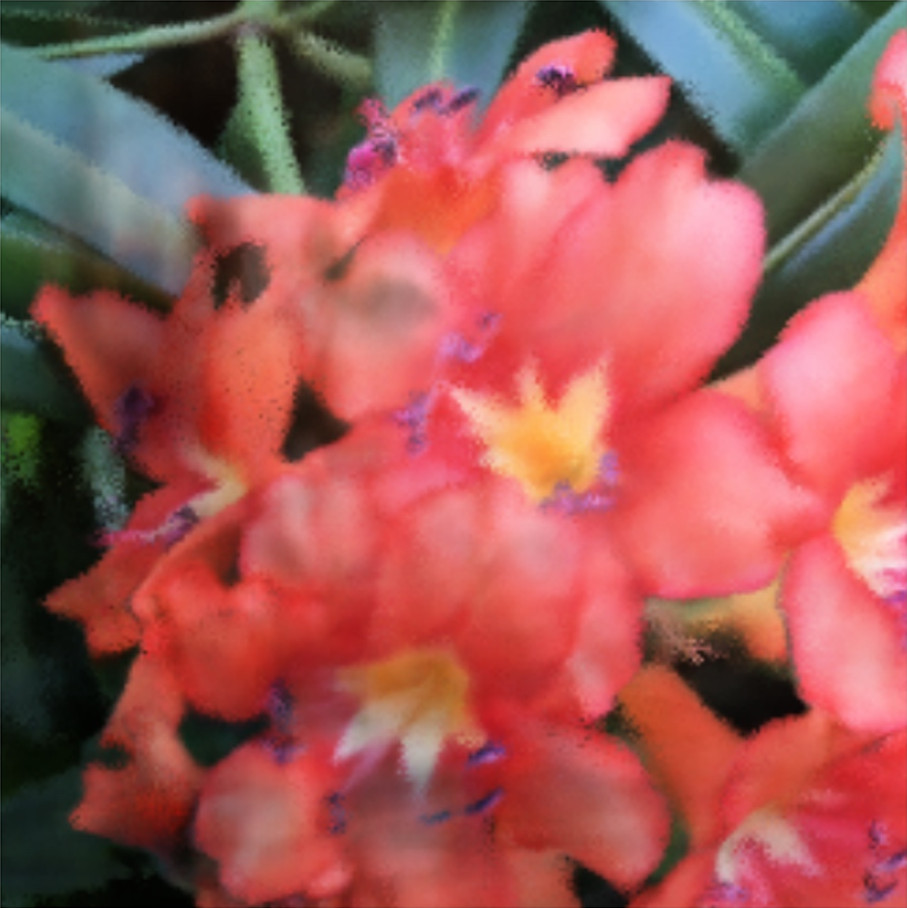} &
            \includegraphics[width=2.7cm]{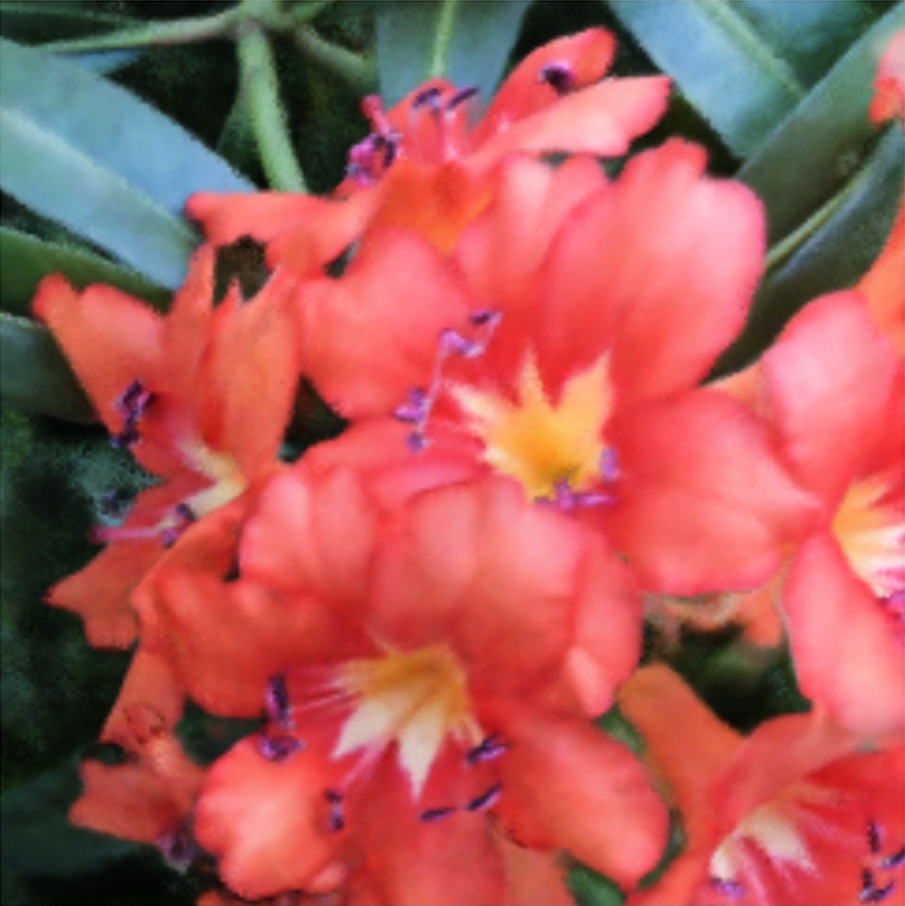} &
            \includegraphics[width=2.7cm]{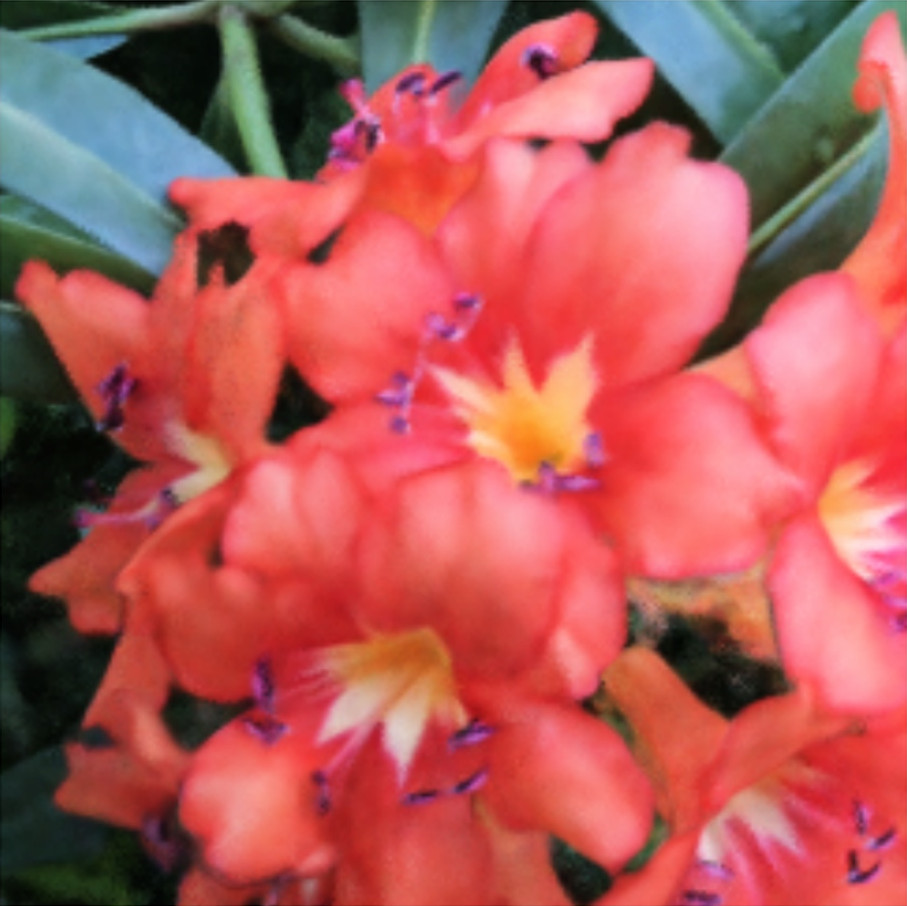} &
            \includegraphics[width=2.7cm]{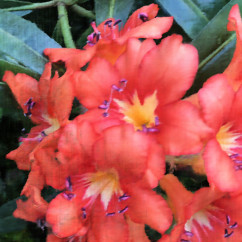} &
            \includegraphics[width=2.7cm]{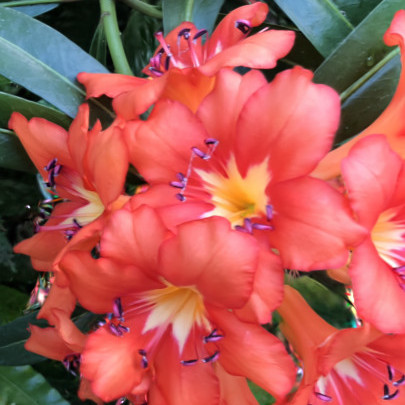} \\
         \rotatebox{90}{\makebox[2.5cm]{trex}} &
            \includegraphics[width=2.7cm]{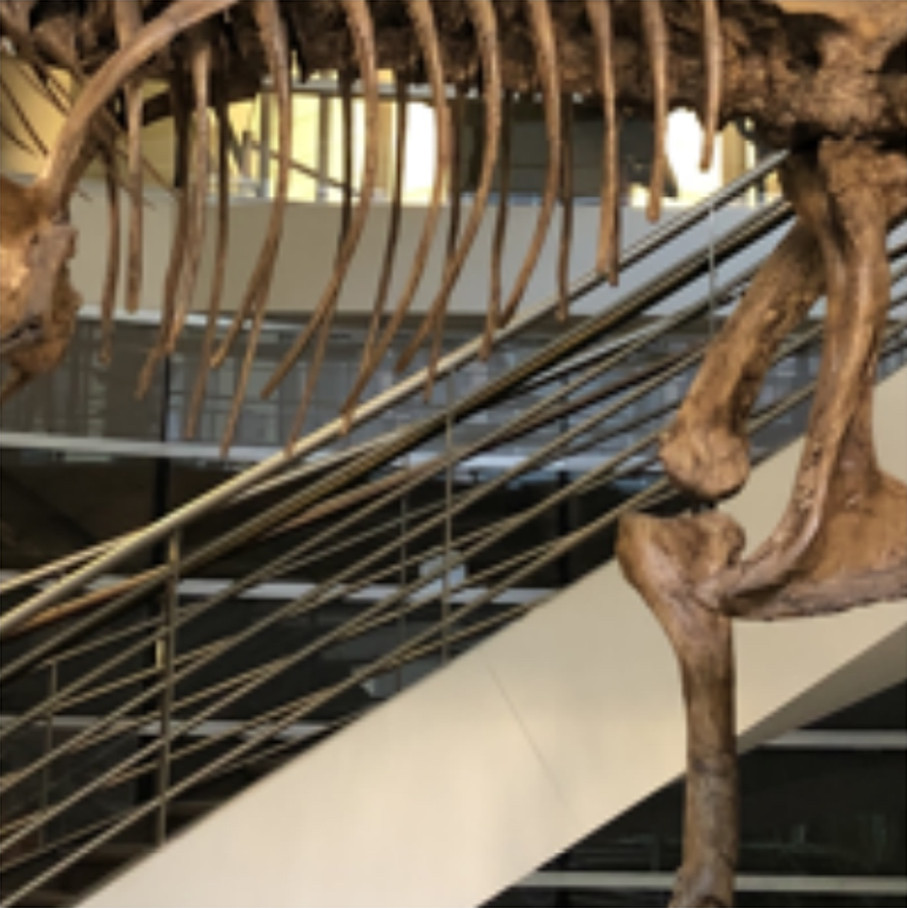} &
            \includegraphics[width=2.7cm]{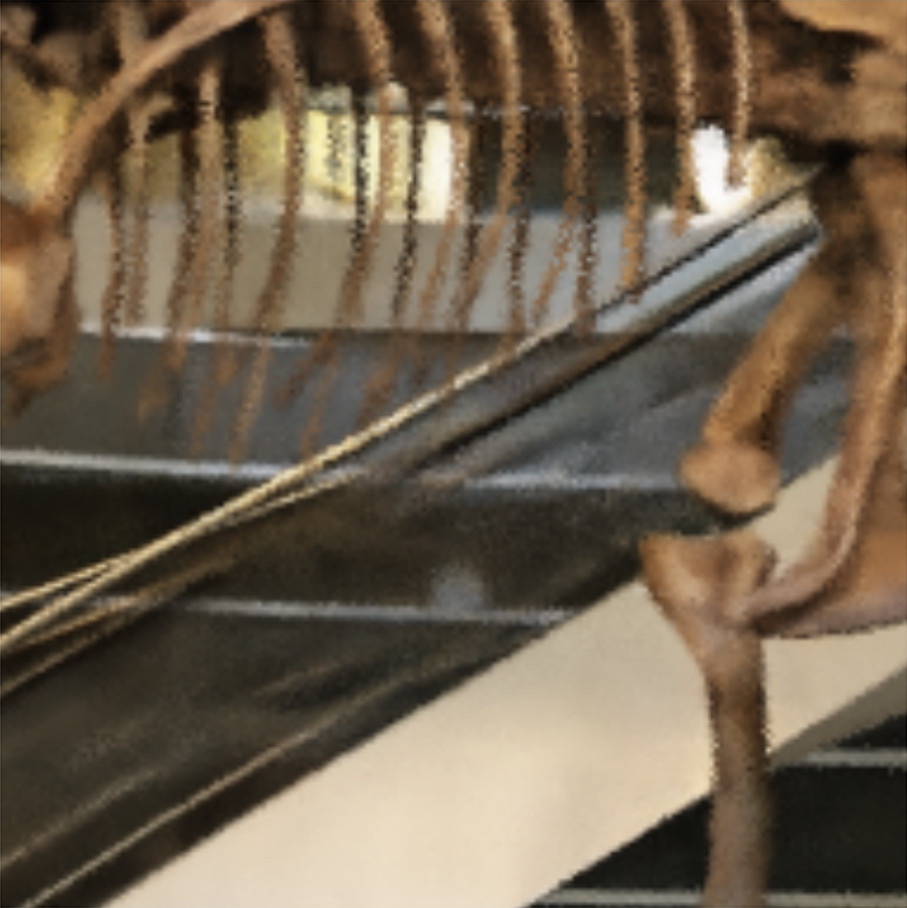} &
            \includegraphics[width=2.7cm]{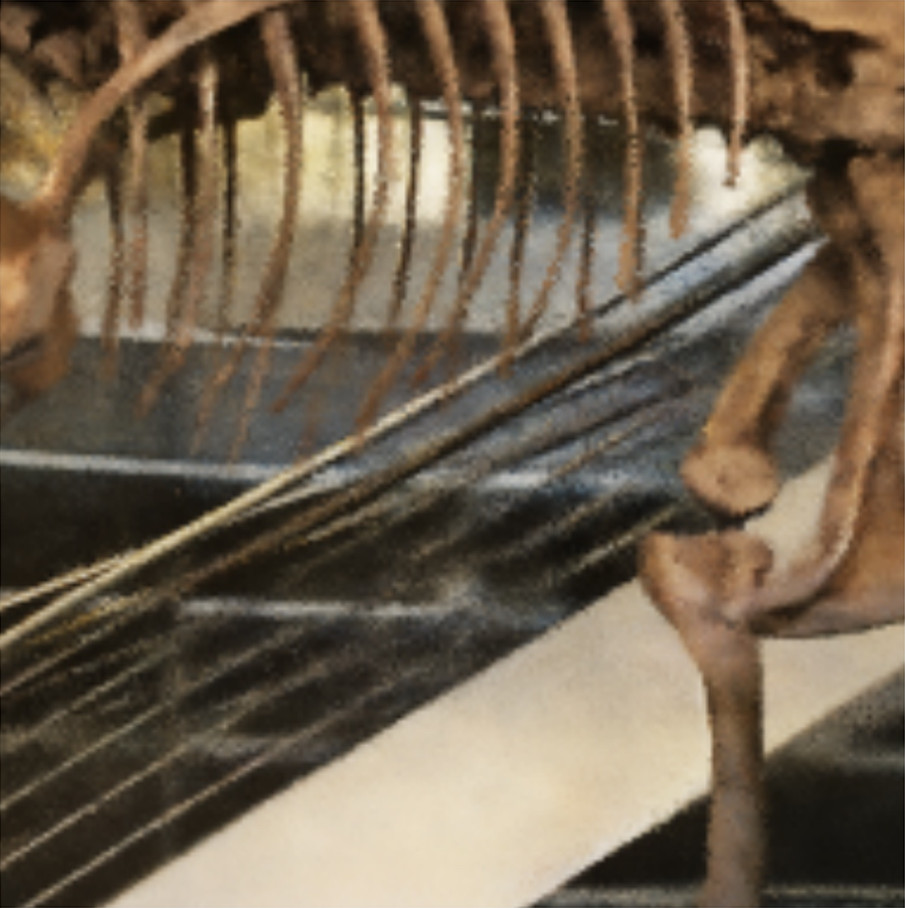} &
            \includegraphics[width=2.7cm]{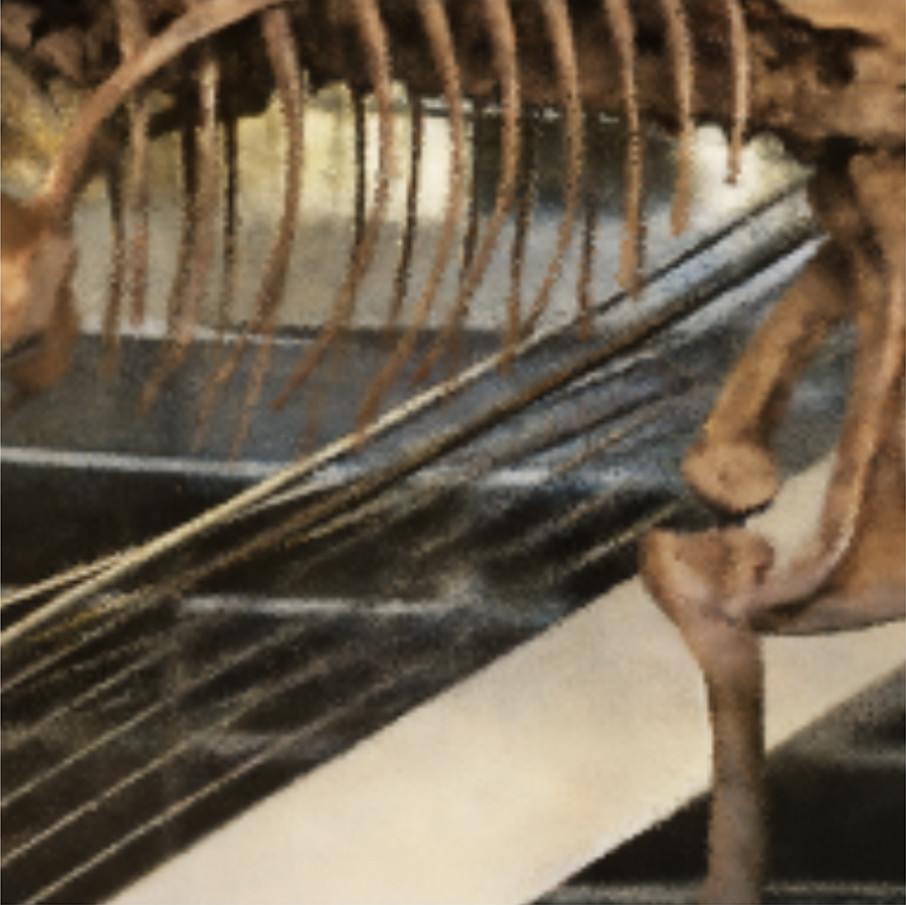} &
            \includegraphics[width=2.7cm]{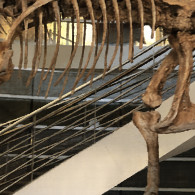} &
            \includegraphics[width=2.7cm]{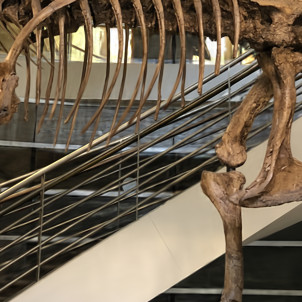} \\
    \end{tabular}
    }
    \caption{
    Qualitative results for pose-free reconstruction and NVS on the \emph{LLFF} dataset~\cite{mildenhall2019llff}.
    Our method achieves superior reconstruction quality compared to previous techniques.
    Notably, BARF, GARF and MRHE struggle to capture fine details, as exemplified by the banister in the trex scene.
    Additionally, their rendered images often exhibit blurriness.
    While JRTF appears sharper at first glance, it suffers from significant pixelation artifacts, particularly evident in the leaves of the flower scene.
    In contrast, our method produces crisp and artifact-free renderings.
    }
    \label{fig:llff-qualitative-comparison}
\end{figure*}

\subsection{Experimental Setup}
We compare against several previous works~\cite{bian2023nope,lin2021barf} as well as a concurrent similar approach~\cite{fu2023cf3dgs}.

\PAR{Datasets.}
We evaluate our method on \emph{LLFF}, \emph{Replica}, and \emph{Tanks and Temples}.
\emph{LLFF}~\cite{mildenhall2019llff} is a set of 8 forward-facing real scenes captured with a handheld smartphone camera.
The ground-truth poses in this dataset are estimated with COLMAP~\cite{schonberger2016colmap}.
As in previous works, we reserve the last 10\% of images for the NVS test set~\cite{lin2021barf}.

\emph{Replica}~\cite{replica19arxiv} contains high-quality renderings of indoor scenes.
In particular, we evaluate our method on the trajectories generated by~\cite{sucar2021imap}, which correspond to 5 offices and 3 apartment rooms.
We subsample the trajectories to 400 frames and use every eighth frame for testing.

\emph{Tanks and Temples}~\cite{knapitsch2017tandt} consists of several real-word scenes captured with a high-resolution camera.
Similar to previous work~\cite{bian2023nope}, we take every eighth image for testing and train on the remaining images.
We use the camera poses estimated by COLMAP as ground truth.
Compared to \emph{LLFF}, \emph{Tanks and Temples} contains much larger scenes with longer and more complex trajectories.

\PAR{Metrics.}
We evaluate our method on the tasks of novel view synthesis (NVS) and camera pose estimation.
For NVS, we report peak-signal-to-noise ratio (PSNR), structural similarity measure (SSIM), and learned perceptual image patch similarity (LPIPS), as commonly done in previous works~\cite{bian2023nope,lin2021barf,cheng2024JointTensoRF,chng2022garf}.
For pose estimation, we use two slightly different sets of metrics.
For \emph{pose refinement}, we use the absolute rotation and translation error between a pose $\mat{T}_i$ and the corresponding ground-truth pose $\mat{T}_i^*$.
For \emph{pose estimation}, we compute the relative rotation and translation error (\RPEr{} and \RPEt, respectively), as well as the absolute trajectory error (ATE).
For all methods and all tasks, we compute pose estimation metrics after aligning the predicted and ground-truth trajectories using Procrustes alignment.

\PAR{Implementation Details.}
We integrate the gradient computation of the camera parameters into the Gaussian Splatting CUDA kernel and implemented the remaining parts of our method in PyTorch.

\PAR{Pose Estimation.}
We optimize the reference models with poses estimated by COLMAP~\cite{schonberger2016colmap}, using our anisotropy loss and adaptive opacity thresholding.
We set $r=10$ and $N_{\text{tgt}}=256'000$.

\PAR{Joint Reconstruction and Pose Refinement.}
For the pose refinement task, we initialize the point cloud by unprojecting points from the depth map of an off-the-shelf monocular depth estimator and treating them as Gaussians as described in~\cref{sec:method:pose-refinement}.
In order to be comparable to previous works we use DPT~\cite{ranftl2021dpt}, in particular the same model and weights as NopeNeRF~\cite{bian2023nope}.
Note that we use depth only for initialization; we do not use ground-truth depth at any point.
We start the camera learning rate at $10^{-2}$ and decay to $10^{-4}$, using a cosine decay schedule.
In the anisotropy regularizer, we set the maximum ratio to $r=10$.
We adopt the majority of hyperparameter settings from the original 3D Gaussian Splatting paper~\cite{kerbl3Dgaussians}, with the following exceptions:
We start with a positional learning rate of $1.6\times10^{-2}$, decaying exponentially to $1.6\times10^{-4}$.
Contrary to the original implementation, the learning rate is not scaled by the scene extent, since we have no (reliable) information on poses.

Similar to previous work~\cite{lin2021barf,chng2022garf,cheng2024JointTensoRF,bian2023nope}, we perform test-time optimization of camera poses in order to minimize the influence of suboptimal pose estimation on NVS performance.
We optimize the pose of each test frame for up to 200 steps using the same hyperparameters as for the relative pose estimation during the online phase.

For the comparison to pose-free methods on video clips, we estimate an initial trajectory using a simple per-frame optimization scheme.
From each frame, we unproject $50'000$ points and optimize them to fit the frame for 100 steps.
We optimize the relative pose between consecutive frames for 200 steps, using a learning rate of $10^{-3}$ decaying to $10^{-4}$ using a cosine schedule.
This scheme is very similar to the initial phase of CF-3DGS~\cite{fu2023cf3dgs}.
However, our method requires much less time to estimate the relative poses due to our efficient CUDA integration.
Additionally, the main part of our method is independent of this simple per-frame scheme; our optimization scheme works based on any initialization with sufficiently small pose errors.
For example, we are able to optimize noisy poses of subsampled trajectories of the Replica dataset~\cite{replica19arxiv} (see~\cref{ssec:replica}), with substantial camera motion between subsequent frames.
In contrast to this, CF-3DGS explicitly proposes an iterative expansion scheme.

We conduct all experiments on a single RTX 3090 GPU.

\subsection{Camera pose estimation from trained models}
\label{sec:pose-estimation}

\begin{figure}
  \centering
  \begin{minipage}{0.485\linewidth}
    \centering
    \includegraphics[width=\linewidth]{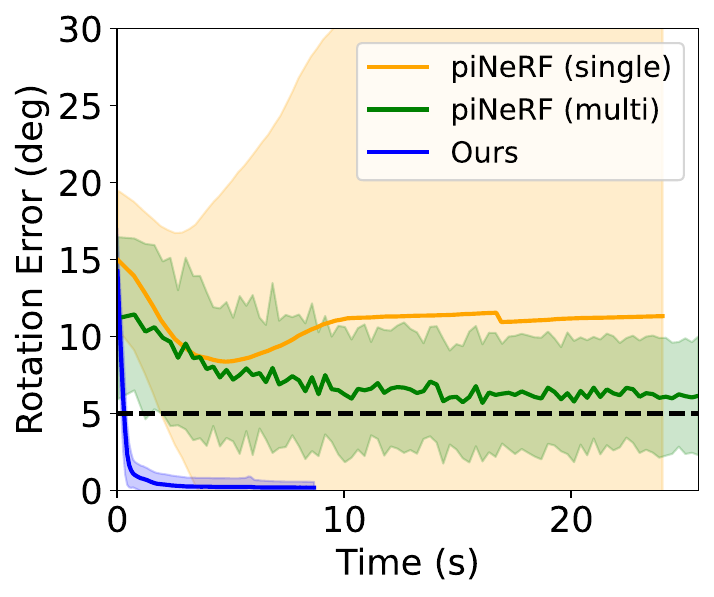}
  \end{minipage}
  \hfill
  \begin{minipage}{0.493\linewidth}
    \centering
    \includegraphics[width=\linewidth]{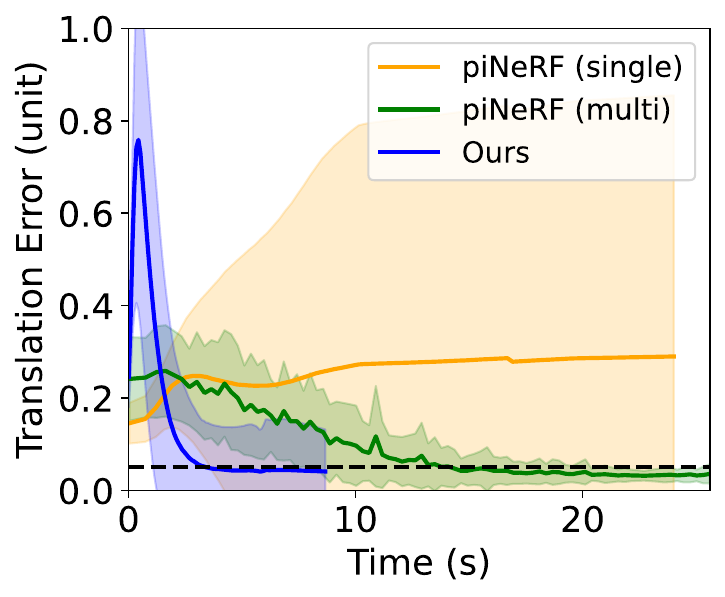}
  \end{minipage}
  \caption{
    Convergence behavior of our method compared to piNeRF~\cite{lin2023icra:pnerf} on a subset of the \emph{LLFF} dataset, \ie fern, flower, fortress, horns and room.
    The dashed black line represents the thresholds of 5deg for rotation and 0.05 for translation.
    The colored bold line is the mean of the error over all scenes.
    Moreover, the +/- std is shown as the colored area around the mean line.
    Our method uses only a single pose hypothesis, but converges much faster than even the multi-hypothesis baseline.
    }
    \label{fig:campose-convergence}
\end{figure}

We first investigate the convergence speed of our method on the task of 6-DoF camera pose estimation from monocular images.
Given a trained model of a scene, a single RGB image, and an initial pose estimate, we want to find the correct position and orientation of the camera corresponding to the image.
We follow previous work~\cite{lin2023icra:pnerf} and generate noisy poses by rotating the camera pose sequentially around each axis, sampling rotational noise uniformly from the range $[-15^\circ, 15^\circ]$ and translating it along the world axes by a random offset sampled from $[-0.15, 0.15]$ units.
We optimize poses for up to 1000 steps, although most trials converge much earlier.
Due to our efficient CUDA implementation, this process takes just a couple of seconds on an NVIDIA RTX 3090.
We evaluate our method on real-world scenes from the LLFF dataset~\cite{mildenhall2019llff} and compare against piNeRF~\cite{lin2023icra:pnerf}, a recent approach which uses a multi-hypothesis optimization strategy based on Instant-NGP~\cite{mueller2022instant}.
A visual comparison is shown in~\cref{fig:campose-convergence}, and quantitative results are shown in~\cref{tab:llff-campose}.
Compared to the piNeRF baseline, our approach converges much faster on the rotation error, even though piNeRF uses a highly performant hashgrid representation and optimizes multiple pose hypotheses in parallel.
We hypothesize that the initial increase in translational error is due to the fact that we optimize camera poses as elements of $\SEthree$; this leads to a coupling between camera position and camera orientation, as also noted in~\cite{lin2023icra:pnerf}, which optimizes poses in $\TxSOthree$.
However, our method still shows rapid convergence after the initial spike.
We leave a thorough investigation of the effect of different Lie Group parametrizations and potential multi-hypothesis setups of our method for future work.

\begin{table}
    \caption{
        Mean absolute rotation and translation error for camera pose estimation on LLFF~\cite{mildenhall2019llff}.
    }
    \label{tab:llff-campose}
    \centering
    \begin{adjustbox}{max width=\textwidth}
    \begin{tabular}{l cc cc}
    \toprule
    Scene & Pos & Rot & Rot@5 & Pos@0.05 \\ %
    \midrule
    Fern     &  0.013 & 0.009 & 1.000 & 1.000 \\ %
    Flower   &  0.024 & 0.035 & 1.000 & 1.000 \\ %
    Fortress &  0.146 & 0.868 & 0.933 & 0.800 \\ %
    Horns    &  0.020 & 0.015 & 1.000 & 0.975 \\ %
    Room     &  0.004 & 0.000 & 1.000 & 1.000 \\ %
    \midrule
    Mean     &  0.041 & 0.185 & 0.987 & 0.955 \\ %
    \midrule\midrule
    \cite{lin2023icra:pnerf} single & 0.311 & 11.054 & 0.680 & 0.656 \\ %
    \cite{lin2023icra:pnerf} multi  & 0.038 & 5.428 & 0.808 & 0.760 \\ %
    \bottomrule
    \end{tabular}
    \end{adjustbox}
\end{table}

\subsection{Joint Reconstruction and Pose Refinement.}
\label{sec:pose-refinement}

\PARbegin{Real-World Scenes.}
We investigate the performance of our method on LLFF, a set of forward-facing real-world scenes captured with a handheld smartphone camera.
In this setting, we assume \emph{unknown} camera poses and initialize all camera poses to identity.
We compare against several state-of-the-art methods for joint reconstruction and pose estimation~\cite{lin2021barf,cheng2024JointTensoRF,Heo2023RobustCP} in~\cref{tab:exp:llff-reconstruction}. Qualitative results are shown in Fig.~\ref{fig:llff-qualitative-comparison}.
Our method outperforms all other methods in visual quality, and performs well in terms of camera registration accuracy, while being several times faster to optimize.

\begin{table}
    \vspace{0.2cm}
    \centering
    \caption{
        Novel view synthesis and pose estimation results on LLFF~\cite{mildenhall2019llff}, training/optimization time of one scene.
    }
    \label{tab:exp:llff-reconstruction}
    \begin{adjustbox}{max width=\linewidth}
    \begin{tabular}{l ccc cc c}
        \toprule
        \multirow{2}{*}{Method} & \multicolumn{3}{c}{Novel View Synthesis} & \multicolumn{2}{c}{Pose Estimation} & \multirow{2}{*}{Time} \\
        \cmidrule(lr){2-4}\cmidrule(lr){5-6}
         & \thPSNR & \thSSIM & \thLPIPS & \thROT & \thPOS &  \\
        \midrule
        BARF~\cite{lin2021barf}
            & 23.09 & 0.678 & 0.275 & 1.580 & 0.721 & 300m \\
        GARF~\cite{chng2022garf}
            & 24.55 & 0.745 & 0.216 & \textbf{0.280} & 0.269 & 600m\\
        MRHE~\cite{Heo2023RobustCP}
            & 24.79 & 0.772 & 0.197 & 0.384 & \textbf{0.258} & 30m\\
        JTRF~\cite{cheng2024JointTensoRF}
            & \textbf{25.27} & 0.827 & - & 0.709 & 0.325 & 180m \\
        \midrule
        Ours
            & 25.19 & \textbf{0.838} & \textbf{0.123} & 1.529 & 0.314 & \textbf{10m} \\
        \bottomrule
    \end{tabular}
    \end{adjustbox}
\end{table}

\begin{figure}
    \centering
    \begin{minipage}{0.5\linewidth}
        \includegraphics[width=\textwidth]{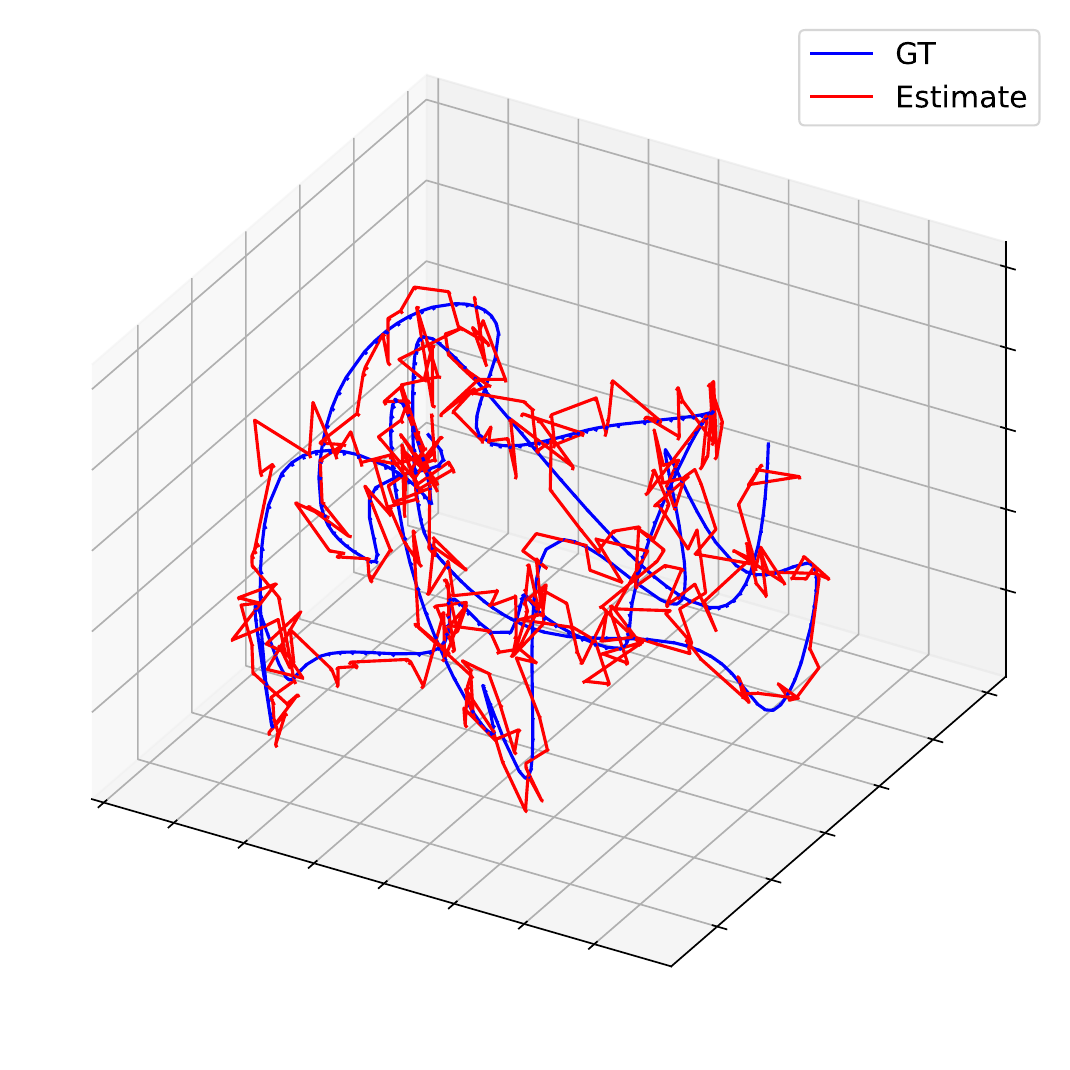}
    \end{minipage}\hfill
    \begin{minipage}{0.5\linewidth}
        \includegraphics[width=\textwidth]{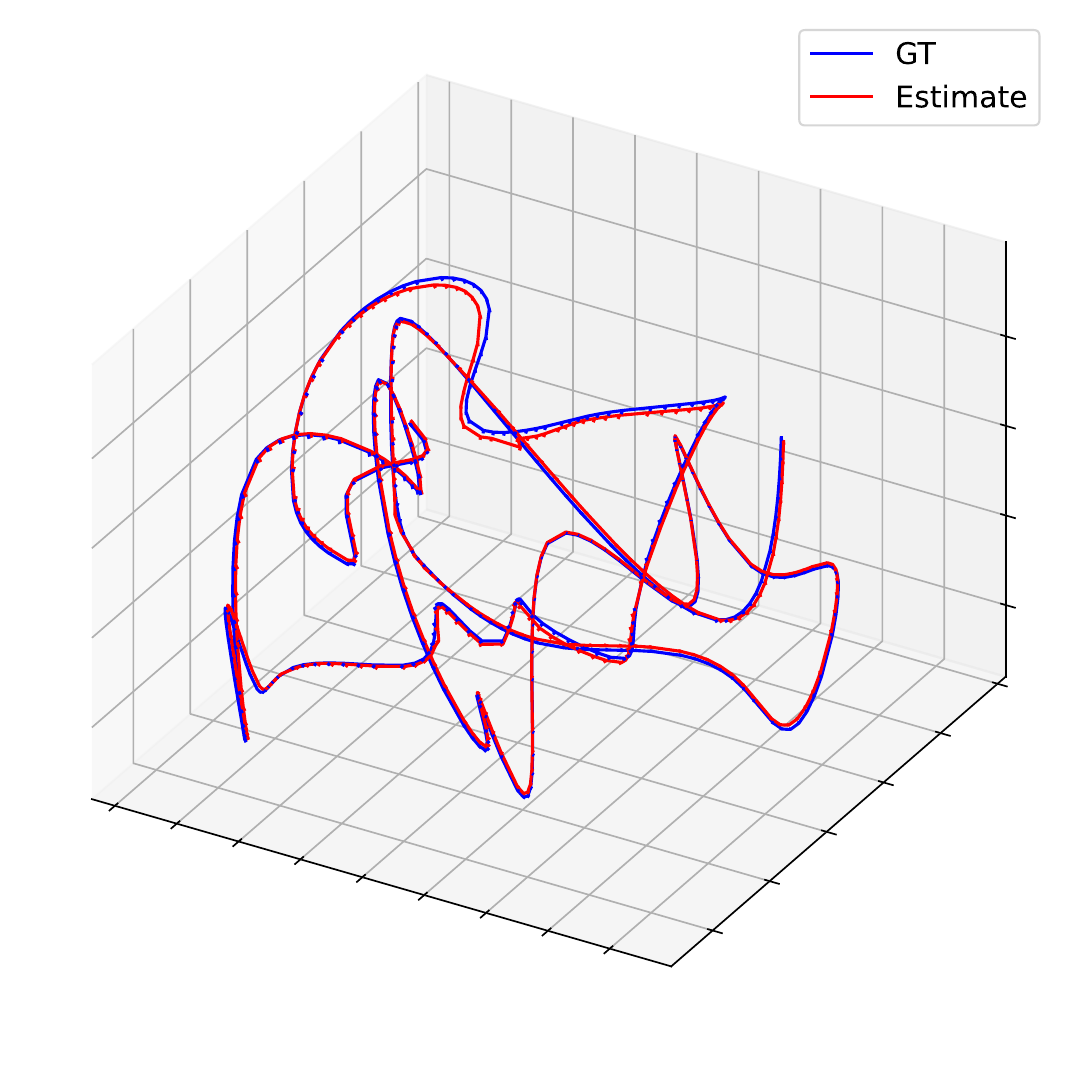}
    \end{minipage}%
    \caption{Qualitative results on \texttt{office0} from \emph{Replica}~\cite{replica19arxiv}. Left: trajectory at the beginning of optimization. Right: trajectory after optimization. Processing takes around 30 minutes on a single RTX 3090.}
    \label{fig:replica-results}
\end{figure}

\PAR{Complex Trajectories.}
\label{ssec:replica}
We evaluate our method on the \emph{Replica} dataset~\cite{replica19arxiv}, using trajectories generated by random walks~\cite{sucar2021imap}.
\emph{Replica} contains 8 room-scale scenes with challenging geometry, lighting, and camera trajectories.
We subsample the trajectories by a factor of 5, and use every eight frame of that for testing.
In total, that yields 350 training views and 50 test views for each of the 8 scenes.
Here, we perturb the poses by sampling gaussian noise in the tangent space with a standard deviation of 0.05.
We show qualitative results in~\cref{fig:replica-results} and quantitative results for NVS and camera pose estimation in~\cref{tab:replica-results}.

\subsection{Reconstruction without Pose Information}
\label{ssec:comparison-sota}

We evaluate NVS performance and pose estimation on a subset of the \emph{Tanks and Temples} benchmark, as proposed in ~\cite{bian2023nope}.
We compare our method against previous NeRF-based approaches that can be trained without known poses, in particular Nope-NeRF~\cite{bian2023nope} and BARF~\cite{lin2021barf}.
In addition, we compare to CF-3DGS~\cite{fu2023cf3dgs}, a concurrent work that also uses Gaussian Splatting for this task.
For all methods, pose estimation performance is evaluated after Procrustes alignment.

We report the results in \cref{tab:exp:comparison-tandt}.
Compared to the NeRF baselines, our method achieves consistently better novel view quality, and a similar visual fidelity as CF-3DGS.
Notably, our method is more than 40$\times$ faster than the NeRF baselines, and about 4$\times$ faster than CF-3DGS (\ie, $20$+~hrs/scene for Nope-NeRF, around 2 hrs/scene for CF-3DGS, and less than 30 minutes/scene for our method).
Again, our method achieves comparable results to state-of-the-art methods, with a fraction of the runtime.

\begin{table}
    \vspace{0.2cm}
    \caption{
        Novel view synthesis and pose estimation results on Replica~\cite{replica19arxiv}.
    }
    \label{tab:replica-results}
    \centering
    \begin{adjustbox}{max width=\linewidth}
    \begin{tabular}{l ccc ccc}
    \toprule
    \multirow{2}{*}{Scene} & \multicolumn{3}{c}{Novel View Synthesis} & \multicolumn{3}{c}{Pose Estimation} \\
    \cmidrule(lr){2-4}\cmidrule(lr){5-7}
             & \thPSNR & \thSSIM & \thLPIPS & \thRPEt & \thRPEr & ATE \\
    \midrule
    Room0    & 28.169 & 0.858 & 0.180 & 1.968 & 0.364 & 0.056 \\
    Room1    & 30.906 & 0.899 & 0.170 & 2.316 & 0.493 & 0.045 \\
    Room2    & 31.535 & 0.928 & 0.161 & 1.368 & 0.357 & 0.034 \\
    Office0  & 42.354 & 0.982 & 0.078 & 0.173 & 0.047 & 0.009 \\
    Office1  & 40.510 & 0.970 & 0.149 & 0.652 & 0.176 & 0.041 \\
    Office2  & 33.245 & 0.938 & 0.193 & 0.687 & 0.256 & 0.061 \\
    Office3  & 35.478 & 0.957 & 0.128 & 0.573 & 0.093 & 0.034 \\
    Office4  & 32.487 & 0.938 & 0.177 & 2.458 & 0.479 & 0.125 \\
    \midrule
    Mean     & 34.335 & 0.934 & 0.154 & 1.274 & 0.283 & 0.051 \\
    \bottomrule
    \end{tabular}
    \end{adjustbox}
\end{table}

\begin{table}
    \vspace{0.2cm}
    \centering
    \caption{
        Comparison on Tanks and Temples~\cite{knapitsch2017tandt} against state-of-the-art methods for joint 3D reconstruction and pose estimation. $^*$ denotes concurrent work.
    }
    \label{tab:exp:comparison-tandt}
    \begin{adjustbox}{max width=\linewidth}
    \begin{tabular}{l ccc ccc }
        \toprule
        \multirow{2}{*}{Method} & \multicolumn{3}{c}{Novel View Synthesis} & \multicolumn{3}{c}{Pose Estimation} \\
        \cmidrule(lr){2-4}\cmidrule(lr){5-7}
         & \thPSNR & \thSSIM & \thLPIPS & \thRPEt & \thRPEr & \thATE \\
         \midrule
         Nerf-\,-~\cite{wang2021nerfmm} & 22.50 & 0.59 & 0.54 & 1.735 & 0.477 & 0.123 \\
         SCNerf~\cite{jeong2021scnerf} & 23.76 & 0.65 & 0.48 & 1.890 & 0.489 & 0.129 \\
         BARF~\cite{lin2021barf} & 23.42 & 0.61 & 0.54 & 1.046 & 0.441 & 0.078 \\
         Nope-Nerf~\cite{bian2023nope} & 26.34 & 0.74 & 0.39 & 0.080 & 0.038 & 0.006 \\
         CF-3DGS$^*$~\cite{fu2023cf3dgs} & 31.28 & 0.93 & 0.09 & 0.041 & 0.069 & 0.004 \\
         \midrule
         Ours & 31.24 & 0.92 & 0.12 & 0.075 & 0.069 & 0.009 \\
         \bottomrule
    \end{tabular}
    \end{adjustbox}
\end{table}

\subsection{Ablations}
\label{sec:ablations}

\PARbegin{Comparison against 3DGS+COLMAP.}
We report the results of vanilla Gaussian Splatting with poses estimated by COLMAP in~\cref{tab:ablation_vanilla_3dgs}.
The visual quality of our reconstruction is comparable to splatting using SfM poses, in some cases even better (\eg, the \emph{Family} and \emph{Horse} sequences of \emph{Tanks and Temples}, see~\cref{tab:ablation_vanilla_3dgs}).
This is likely due to a suboptimal reconstruction by COLMAP, leading to slightly wrong camera poses or misplacement of initial 3D points for the initialization of 3DGS.
Similarly, we observe a higher visual quality for the forward-facing \emph{LLFF} scenes.
The high visual quality and low pose error make our method interesting for postprocessing of, \eg, noisy SLAM trajectories.

\begin{table*}
\vspace{0.15cm}
    \label{tab:ablation_vanilla_3dgs}
    \caption{
        Comparison between our method, trained without pose information, and Gaussian Splatting using poses estimated with COLMAP.
        Left: Results on Tanks and Temples~\cite{knapitsch2017tandt}.
        Right: Results on LLFF~\cite{mildenhall2019llff}.
    }
    \centering
    \begin{minipage}{.49\textwidth}
    \begin{adjustbox}{max width=\textwidth}
    \begin{tabular}{l ccc ccc}
        \toprule
        \multirow{2}{*}{Scene} & \multicolumn{3}{c}{Ours} & \multicolumn{3}{c}{3DGS+COLMAP} \\
        \cmidrule(lr){2-4}\cmidrule(lr){5-7}
        & \thPSNR & \thSSIM & \thLPIPS & \thPSNR & \thSSIM & \thLPIPS\\
        \midrule
        Ballroom & 32.73 & 0.96 & 0.05 &   34.66 & 0.97 & 0.03 \\
        Barn     & 31.21 & 0.89 & 0.15 &   32.11 & 0.96 & 0.07 \\
        Church   & 28.64 & 0.89 & 0.16 &   30.00 & 0.94 & 0.09 \\
        Family   & 32.02 & 0.94 & 0.10 &   28.43 & 0.93 & 0.12 \\
        Francis  & 32.66 & 0.91 & 0.17 &   32.23 & 0.92 & 0.16 \\
        Horse    & 33.20 & 0.96 & 0.07 &   21.60 & 0.80 & 0.22 \\
        Ignatius & 28.56 & 0.87 & 0.18 &   30.27 & 0.93 & 0.09 \\
        Museum   & 30.86 & 0.92 & 0.11 &   34.73 & 0.97 & 0.05 \\
        \midrule
        Mean     & \textbf{31.24} & 0.92 & 0.12 &   30.50 & \textbf{0.93} & \textbf{0.10} \\
         \bottomrule
    \end{tabular}
    \end{adjustbox}
    \end{minipage}\hfill
    \begin{minipage}{0.49\textwidth}
    \begin{adjustbox}{max width=\textwidth}
    \begin{tabular}{l ccc ccc}
        \toprule
        \multirow{2}{*}{Scene} & \multicolumn{3}{c}{Ours} & \multicolumn{3}{c}{3DGS+COLMAP} \\
        \cmidrule(lr){2-4}\cmidrule(lr){5-7}
        & \thPSNR & \thSSIM & \thLPIPS & \thPSNR & \thSSIM & \thLPIPS\\
        \midrule
        \makebox[0pt][l]{Fern}\phantom{Ballroom}     & 26.313 & 0.867 & 0.111 & 23.322 & 0.790 & 0.232 \\
        Flower   & 25.297 & 0.834 & 0.120 & 26.802 & 0.834 & 0.226 \\
        Fortress & 30.215 & 0.924 & 0.080 & 29.530 & 0.872 & 0.180 \\
        Horns    & 22.210 & 0.880 & 0.123 & 24.464 & 0.837 & 0.237 \\
        Leaves   & 18.957 & 0.686 & 0.217 & 17.480 & 0.586 & 0.305 \\
        Orchids  & 18.201 & 0.627 & 0.205 & 18.780 & 0.624 & 0.259 \\
        Room     & 35.354 & 0.980 & 0.040 & 32.272 & 0.942 & 0.114 \\
        Trex     & 25.007 & 0.909 & 0.084 & 23.828 & 0.869 & 0.227 \\
        \midrule
        Mean     & \textbf{25.194} & \textbf{0.838} & \textbf{0.123} & 24.560 & 0.794 & 0.223 \\
         \bottomrule
    \end{tabular}
    \end{adjustbox}
    \end{minipage}
\end{table*}

\section{Discussion}
\label{sec:discussion}

We have proposed a novel approach that integrates camera pose estimation into 3D Gaussian Splatting by extending the high-performance CUDA rendering kernel.
We verified the efficacy of our method for camera pose estimation and joint reconstruction and pose refinement on several well-known benchmark datasets, achieving state-of-the art results in novel view synthesis quality and pose estimation accuracy, while being several times faster than competing methods. 

3D Gaussian Splatting approaches are rapidly gaining traction in many application areas where NeRFs have previously been successful, due to their extremely fast differentiable rendering capabilities. Our proposed approach enables those approaches to perform more robustly with less requirements on accurate pose initialization, while preserving the supreme runtime speed that makes 3DGS attractive.

\textbf{Acknowledgements.}
Christian Schmidt was funded by BMBF project bridgingAI (16DHBKI023).
Jens Piekenbrink was funded by Bosch Research as part of the Bosch-RWTH Lighthouse collaboration “Context Understanding for Autonomous Systems”.
The authors would like to thank Alexander Hermans for his valuable feedback and discussions.

\bibliographystyle{IEEEtran}
\bibliography{main}

\end{document}